\documentstyle[11pt,aclap,psfig]{article}

\title{An Efficient Distribution of Labor in a Two Stage Robust Interpretation
Process}
\author{Carolyn Penstein Ros\'{e}\\
Carnegie Mellon University\\
Baker Hall 135F\\
Pittsburgh, PA 15213\\
cprose@cs.cmu.edu \And
Alon Lavie \\
Carnegie Mellon University \\
Center for Machine Translation \\
Pittsburgh, PA 15213 \\
alavie@cs.cmu.edu}

\begin{document}
\maketitle\maketitle

\bibliographystyle{fullname}

\begin{abstract}

Although Minimum Distance Parsing (MDP) offers a theoretically attractive solution 
to the problem of extragrammaticality, it is often computationally infeasible 
in large scale practical applications.  In this paper we present an 
alternative approach where the labor is distributed between a more 
restrictive partial parser and a repair module. 
Though two stage approaches have grown in popularity in recent years because
of their efficiency, they have done so at the cost of requiring hand coded
repair heuristics \cite{uteger,danieli}.  In contrast, our two stage approach 
does not require any hand coded knowledge sources dedicated to repair,
thus making it possible to achieve a similar run time advantage over MDP
without losing the quality of domain independence.
\end{abstract}

\section{Introduction}

The correct interpretation of spontaneous spoken language poses challenges 
that continue to fall outside of the reach of state-of-the-art technology.  
The first essential task of a natural language interface is to map the user's
utterance onto some meaning representation which can then be used for
further processing.  The three biggest challenges that continue to stand
in the way of accomplishing even this most basic task are extragrammaticality,
ambiguity, and speech recognition errors.  In this paper we address the issue
of how to handle the problem of extragrammaticality efficiently, where
extragrammaticality is defined as any deviation of an input string from 
the coverage of a given system's parsing grammar.  We demonstrate the
superiority of our approach by comparing performance between it and
a set of alternative approaches in terms of parse time
and parse quality over the same previously unseen test corpus.

The approach presented in this paper is the completely automatic portion
of the ROSE\footnote{ROSE is pronounced ros\'{e}, like the wine.} approach.  
ROSE, RObustness with Structural Evolution, repairs  extragrammatical input in 
two phases.  The first phase, Repair Hypothesis Formation, is responsible 
for assembling a set of hypotheses about the meaning of the ungrammatical 
utterance. This phase is itself divided into two stages, Partial Parsing
and Combination. 
A restricted version of Lavie's GLR* parser \cite{lavie,lt} is used to obtain 
an analysis of
islands of the speaker's sentence in cases where it is not possible to
obtain an analysis for the entire sentence.
In the Combination stage, the fragments from the partial parse are assembled 
into a set of alternative meaning representation hypotheses.   A genetic 
programming approach is used to search for different ways to combine the 
fragments in order to avoid requiring any hand-crafted repair rules.  In 
ROSE's second phase, Interaction with the User, the system generates a set 
of queries, negotiating with the speaker in order to narrow down to a single 
best meaning representation hypothesis.  In this paper, only the Hypothesis
Formation phase is described and evaluated.  Since repairs beyond those made
possible by the partial parser are performed during the Combination stage,
we refer to the implementation of the Combination stage as the repair module.
Though a set of hypotheses are produced by during the Combination stage,
in the evaluation presented in this paper, only the repair hypothesis
scored by the repair module as best is returned.

The ROSE approach was developed in the context of
the JANUS large-scale multi-lingual machine translation system 
\cite{janus96,janus,janus2}.
Currently, the JANUS system deals with the scheduling domain where
two speakers attempt to
schedule a meeting together over the phone.  The system is composed of
four language independent and domain independent modules including 
speech-recognition, parsing, discourse processing, and generation.  The repair
module described in this paper is similarly language independent and domain
independent, requiring no hand-coded knowledge dedicated to repair.
The evaluations described in this paper were conducted using a grammar with 
approximately 1000 rules and a lexicon with approximately 1000 lexical items.

\section{Alternative Avenues Towards Robustness}

There are a wide range of different approaches to handling the problem of
extragrammaticality, but which way is best?  Three basic avenues exist
whereby the coverage of a natural language understanding system can
be expanded: further development of the parsing grammar, addition of 
flexibility to the parsing algorithm, or addition of a post-processing 
repair stage after the parsing stage.  

It is always possible to add additional 
rules to a parsing grammar in order to expand the coverage, but this approach 
is both time intensive in terms of development and ultimately computationally
expensive at run time since large, cumbersome grammars generate excessive
amounts of ambiguity.  Adding flexibility to the parsing algorithm is
preferable in some respects, particularly in that it reduces the grammar
development burden.  However, it lends itself to the same weakness in terms
of computational expense.  In the extreme case, in a Minimum Distance Parser
(MDP) \cite{lehman,hipp}, any ungrammatical sentence can be mapped onto a 
sentence inside of the 
coverage of the grammar through a series of insertions, deletions, and in some
cases substitutions or transpositions.  

The more flexibility, the better the
coverage in theory, but in realistic large scale systems this approach becomes
computationally intractable.  Current efforts
towards robust interpretation have focused on less powerful partial parsers
\cite{abney,vanord,srin,fedmonpir} and repair approaches 
where the labor is distributed between two or more stages 
\cite{uteger,danieli}.  The purpose of the second stage is
to assemble the pieces of the partial parse produced in the first stage.
In this paper we present a two stage approach composed of a
partial parser followed by a completely automatic repair module.

Though two stage approaches have grown in popularity in recent years because
of their efficiency, they have done so at the cost of requiring hand coded
repair heuristics \cite{uteger,danieli}.  In contrast, the ROSE approach
does not require any hand coded knowledge sources dedicated to repair,
thus making it possible to achieve the benefits of repair
without losing the quality of domain independence.

In this paper, we compare the performance of the two stage ROSE approach with 
MDP.  A parameterized version of Lavie's GLR* parser
\cite{lavie} is used which has been extended to perform a limited version
of MDP in which insertions and deletions are possible, but not transpositions
or substitutions.  We refer to this parameterized MDP parser as LR MDP.
We run LR MDP over the
same test corpus in different settings, demonstrating the 
flexibility/quality/parse time trade off.  With this we demonstrate that the 
two stage ROSE approach, coupling the restricted version of the GLR* parser 
with a post-processing repair stage, achieves 
better translation quality far more efficiently than any flexibility setting
of LR MDP over the same corpus.

\section{MDP versus Two Stage Interpretation}

Efforts towards solving the problem of extragrammaticality
have primarily been in the direction of building flexible parsers.
In principle, Minimum Distance Parsers \cite{lehman,hipp} have the greatest
flexibility.  They fit
an extragrammatical sentence to the parsing grammar through a series of 
insertions, deletions, and transpositions.  Since any string can be mapped onto
any other string through a series of insertions, deletions, and 
transpositions,
this approach makes it possible to repair any sentence.
The underlying assumption behind the MDP approach is that the analysis of
the string that deviates the least from the input string is most likely to
be the best analysis.  Thus, Minimum Distance Parsing appears to be a 
reasonable approach.

In practice, however, Minimum Distance Parsing has only
been used successfully in very small and limited domains.  
Lehman's core grammar, described in \cite{lehman}, has on the order of 300 
rules, and all of the inputs to her system can be assumed to be commands to 
a calendar program.  Hipp's Circuit Fix-It Shop system, described in 
\cite{hipp}, has a vocabulary of only 125 words and a grammar size of only 
500 rules.  Flexible parsing algorithms introduce a great deal of
extra ambiguity.  This in turn may deem certain approaches
impractical for systems of realistic scale.  Therefore, an important question 
one must ask is whether the MDP approach can scale up to a larger system
and/or domain.

An example of a more restrictive parsing algorithm is Lavie's GLR* skipping 
parser described in \cite{lavie}.
GLR* is a parsing system based on Tomita's Generalized LR parsing algorithm
which was designed to be robust to two particular types of 
extra-grammaticality: noise in the input, and limited grammar coverage.  
GLR* attempts to overcome these forms of extra-grammaticality by ignoring the 
unparsable words and fragments and conducting a search for the maximal subset 
of the original input that is covered by the grammar.

The GLR* parser is capable of skipping over any portion of an input utterance
that cannot be incorporated into a grammatical analysis and recover the
analysis of the largest grammatical subset of the utterance.  Partial analyses
for skipped portions of the utterance can also be returned by the parser.  
Thus,
whereas MDP considers insertions and transpositions in addition to deletions,
GLR* only considers deletions.  GLR* can be viewed as a restricted 
form of MDP applied to an efficient non-robust general parsing method.
GLR* can, in most cases, achieve most of the robustness of the more general 
MDP approach while maintaining feasibility, due to efficiency properties of 
the GLR approach and an effective well-guided search.  In the evaluation 
presented in this paper, GLR* has been restricted to skip only initial segments
so that the partial analyses returned are always for contiguous portions of
the sentence.

Because GLR* was designed as an enhancement to the widely used 
standard GLR context-free 
parsing algorithm, grammars, lexicons and other tools developed for the 
standard GLR parser can be used without modification.  GLR* uses the standard 
SLR(0) parsing tables which are compiled in advance from the grammar.  It 
inherits the benefits of GLR in terms of ease of grammar development, and, to 
a large extent, efficiency properties of the parser itself.  In the case that 
an input sentence is completely grammatical, GLR* will normally return the 
exact same parse as the GLR parser.

The weakness of this and other partial 
parsing approaches \cite{abney,vanord,srin,fedmonpir} is that part of 
the original meaning of the utterance may be thrown away with the portion(s) 
of the utterance that are skipped if only the analysis for the largest subset
is returned, or part of the analysis will be missing if the parser only 
attempts to build a partial parse.  These less powerful algorithms trade 
coverage for speed.  The idea is to introduce enough flexibility to gain an 
acceptable level of coverage at an acceptable computational expense.  

The goal behind the two stage approach 
\cite{uteger,danieli} is to increase the coverage possible
at a reasonable computational cost by introducing a post-processing repair
stage, which constructs a complete meaning representation out of the
fragments of a partial parse.  Since the input to the second stage is a 
collection of
partial parses, the additional flexibility that is introduced at this second 
stage can be channeled just to the part of the analysis that the parser does 
not have enough knowledge to handle straightforwardly.  This is unlike the 
MDP approach, where the full amount of 
flexibility is unnecessarily applied to every part of the analysis, even in
completely grammatical sentences.  Therefore, this 
two stage process is a more efficient distribution of labor since the first stage is
highly constrained by the grammar and the results of this first stage are then
used to constrain the search in the second stage.  Additionally, in cases 
where the limited flexibility parser is sufficient, the second stage can be 
entirely bypassed, yielding an even greater savings in time.

\section{The Two Stage Interpretation Process}

The main goal of the two stage ROSE approach is to achieve the ability to 
robustly interpret spontaneous natural language
efficiently in a system at least as large and complex as the 
JANUS multi-lingual machine translation system, which provides the context
for this work.
In this section we describe the division of labor between the Partial Parsing
stage and the Combination stage in the ROSE approach.

\subsection{The Partial Parsing Stage}

\begin{figure}[hbt]
\begin{footnotesize}

\begin{tabbing}

\hspace{.75in} \= \hspace{.20in} \= \\

\> {\bf Sentence:} {\it That wipes out my mornings.} \\ \\

\>\> {\bf Partial Analyses:} \\ \\

\>\> {\bf Chunk1:} that \\ \\

\>\>(\=(ROOT THAT) \\
\>\>\>(TYPE PRONOUN) \\
\>\>\>(FRAME *THAT)) \\ \\

\>\> {\bf Chunk2:} out \\ \\

\>\>((TYPE NEGATIVE) \\
\>\>\>(DEGREE NORMAL) \\
\>\>\>(FRAME *RESPOND)) \\ \\

\>\> {\bf Chunk3:} my \\ \\

\>\>((ROOT I) \\
\>\>\>(TYPE PERSON-POSS) \\
\>\>\>(FRAME *I)) \\ \\

\>\> {\bf Chunk4:} mornings \\ \\

\>\>((TIME-OF-DAY MORNING) \\
\>\>\>(NUMBER PLURAL) \\
\>\>\>(FRAME *SIMPLE-TIME) \\
\>\>\>(SIMPLE-UNIT-NAME TOD)) 

\end{tabbing}
\end{footnotesize}
\caption{\label{parse-example} Parse Example}
\end{figure}

	The first stage in our approach is the Partial 
Parsing stage where the goal is to obtain an 
analysis for islands of the speaker's utterance if it is not possible to 
obtain an analysis for the whole utterance.  This is accomplished with 
a restricted version of Lavie's GLR* parser \cite{lavie,lt} that produces an
analysis for contiguous portions of the input sentence.
See Figure \ref{parse-example} for an example parse.  Here the GLR* parser
attempts to handle the sentence ``That wipes out my mornings.''  The expression
``wipes out'' does not match anything in the parsing grammar.  The grammar
also does not allow time expressions to be modified by possessive pronouns.
So ``my mornings'' also does not parse.  Although the grammar recognizes 
``out'' as a way of expressing a rejection, as in ``Tuesdays are out,'' it does
not allow the time being rejected to follow the ``out''.  However, although the
parser was not able to obtain a complete parse for this sentence, it was
able to extract four chunks.

The chunks are feature structures in which the parser encodes the meaning 
of portions of the user's sentence.  This frame based meaning representation
is called an interlingua because it is language independent.  It is defined
by an interlingua specification, which serves as the primary
symbolic knowledge source used during the Combination stage.
Each frame encodes a concept in the domain.  The set of
frames in the meaning representation are arranged into subsets that are
assigned a particular type.  Each frame is associated with a set of slots.  
The slots represent relationships between feature
structures.  Each slot is associated with a type which determines the set of
possible frames which can be fillers of that slot. 
Though this meaning representation specification is knowledge
that must be encoded by hand, it is knowledge that can be used by all 
aspects of the system, not only the repair module as is the case with
repair rules.  Arguably, any well designed system would have such a
specification to describe its meaning representation.

The four chunks extracted by the parser each encode a different part of the
meaning of the sentence ``That wipes out my mornings.''  The first chunk
represents the meaning of ``that''.  The second one represents the meaning
of ``out''.  Since ``out'' is generally a way of rejecting a meeting time in this
domain, the associated feature structure represents the concept of a 
response that is a rejection.    Since ``wipes'' 
does not match anything in the grammar, this token is left without any
representation among the fragments returned by the parser.  The last two
chunks represent the meaning of ``my'' and ``mornings'' respectively.

The disadvantage of this skipping parser over the MDP approach
is that it does not have the ability to perform some necessary repairs
that the more complicated approach can make.  In this case, for example, it is
unable to determine how these pieces fit together into one coherent parse.
The goal of the Combination stage is to
overcome this limitation efficiently.  Thus, the second stage of the 
interpretation
process is responsible for making the remaining types of repairs.
More flexibility can be introduced in the second stage efficiently since
the search space has already been reduced with the addition of the knowledge
obtained from the partial parse. 

\subsection{The Combination Stage}

The purpose of the Combination stage is to
make the remainder of the types of repairs that
could in principle be done with a minimum distance parser using
insertions, deletions, and transpositions, 
but that cannot be performed with the skipping parser.  
The Combination stage takes as input the partial analyses returned by the
skipping parser.
These chunks are combined into a set of best repair hypotheses.
The hypotheses built during this combination process specify how to build 
meaning representations out of 
the partial analyses produced by the parser that are meant to represent
the meaning of the speaker's whole sentence, rather than just parts.  Since
the meaning representation is compositional, a single, more complete meaning
representation can be built by assembling the meaning representations for the
parts of the sentence.

\begin{figure}[hbt]
\begin{footnotesize}
\begin{tabbing}

\hspace{.05in} \= \\

\> {\bf Ideal Repair Hypothesis:} \\ \\

\> (MY-COMB \hspace{.3in} ;insert arg2 into arg1 in slot \\
\> \hspace{.05in} \= (\= (FRAME *RESPOND) \\
\>                \>\>(DEGREE NORMAL) \\
\>                \>\>(TYPE NEGATIVE)) \hspace{.8in} \= ; arg1 \\
\>\> ((TIME-OF-DAY MORNING) \\
\>\>\>(NUMBER PLURAL)  \\
\>\>\> (FRAME *SIMPLE-TIME) \\
\>\>\> (SIMPLE-UNIT-NAME TOD)) \> ; arg2 \\
\>\>  WHEN) \>\> ; slot 

\end{tabbing}
\begin{tabbing}

\hspace{.05in} \= \\

\>{\bf Ideal Structure:} \\ \\

\>(\=(FRAME *RESPOND) \\
\>\> (DEGREE NORMAL) \\
\>\> (TYPE NEGATIVE) \\
\>\> (WHEN (\=(FRAME *SIMPLE-TIME) \\
\>\>\>        (TIME-OF-DAY MORNING) \\
\>\>\>        (NUMBER PLURAL) \\
\>\>\>        (SIMPLE-UNIT-NAME TOD)))) \\ \\

\> {\bf Gloss:} Mornings are out. 

\end{tabbing}
\end{footnotesize}
\caption{\label{comb-example} Combination Example}
\end{figure}

In this Combination stage, a genetic programming \cite{gen2,gen3}
approach is used to evolve a population of programs that specify how to
build complete meaning representations from the chunks returned from the
parser. The repair module must determine not only which subset of chunks 
returned by the parser to include in the final result, but also how to 
put them together.
For example, the ideal repair hypothesis for the example in Figure
\ref{comb-example} is one that specifies that the temporal expression
should be inserted into the {\tt WHEN} slot in the {\tt *RESPOND} frame.  
The repair process is analogous in some ways 
to fitting pieces of a puzzle into a mold that contains receptacles for 
particular shapes.  In this analogy, the meaning representation specification
acts as the mold with receptacles of different shapes, making it possible to 
compute all of the ways partial analyses can fit together in order to create a 
structure that is legal in this frame based meaning representation.  

Both the skipping 
parsing algorithm and the genetic programming combination algorithm 
are completely domain independent.  
Therefore, the ROSE approach maintains the positive quality
of domain independence that the minimum distance parsing approach has while
avoiding some of the computational expense.

\section{The Genetic Programming Combination Process In-Depth}

Recovery from parser failure is a natural application for genetic programming
\cite{gen2,gen3}.
One can easily conceptualize the process of constructing a meaning 
representation hypothesis as the execution of a computer program
that assembles the set of chunks returned from the parser.
This program would specify the operations required for building larger chunks 
out of smaller chunks and then even larger ones from those.  Because the
programs generated by the genetic search are hierarchical, they
naturally represent the compositional nature of the repair process.  

\subsection{Constructing Alternative Hypotheses}

See Figure 
\ref{comb-example} for an example repair hypotheses.  {\tt MY-COMB} is a simple
function that attempts to insert the second feature structure into some
slot in the first feature structure.  It selects a slot, if a suitable one
can be found, and then instantiates the third parameter to this slot.  In
this case, the {\tt WHEN} slot is selected.  So the feature structure 
corresponding to ``mornings'' is inserted into the {\tt WHEN} slot in the
feature structure corresponding to ``out''.  The result is a feature
structure indicating that ``Mornings are out.''  Though this is not an
exact representation of the speaker's meaning, it is the best that can be
done with the available feature structures\footnote{Note
that part of the expression ``wipes out'' matches a rule
in the grammar that happens to have a similar meaning since ``out''
can be used as a rejection as in ``Tuesday is out.''  If the
expression had been ``out of sight'', which is positive,
both the ROSE approach and MDP would construct the opposite 
meaning from the intended
meaning.  Problems like this can only be dealt with through interaction with
the user to confirm that repaired meanings reflect the speaker's true 
intention.}.  Notice that since the expression
``wipes out'' is foreign to the parsing grammar, and no similar expression
is associated with the same meaning in it, the MDP approach would also
not be able to do better than this since it can only insert and delete in
order to fit the current sentence to the rules in its parsing grammar.  
Additionally, since the time expression
follows ``out'' rather than preceding it as the grammar expects, only MDP with 
transpositions in addition to insertions and deletions would be able to arrive 
at the same result.  Note that the feature structures corresponding to ``my''
and ``that'' are not included in this hypothesis.  The job of the Combination
Mechanism is both to determine which fragments to include as well as how to
combine the selected ones.

In the genetic programming
approach, a population of programs are evolved that specify how to
build complete meaning representations from the chunks returned from the
parser. A complete meaning representation is one that is meant to
represent the meaning of the speaker's whole utterance, rather than
just part.  Partial solutions are evolved through the genetic search
specifying how to build parts of the full meaning representation. 
 Because in the same population there can be programs 
that specify how to build different parts of the meaning representation, 
different parts of the full solution are evolved in parallel, making it 
possible to evolve a complete solution quickly.  

\begin{figure}[htb]
\begin{footnotesize}
\begin{tabbing}

\hspace{.05in} \= \hspace{.02in} \= \\

\>{\bf Hypothesis1:} \\ \\

\>(MY-COMB \\
\> \hspace{.15in} \=  (MY-COMB \\
\>\>\hspace{.15in} \=     (\=(FRAME *RESPOND)) \\
\>\>\>     ((FRAME *SIMPLE-TIME) \\
\>\>\>\>(TIME-OF-DAY MORNING) \\
\>\>\>\>    (NUMBER PLURAL) \\
\>\>\>\> (SIMPLE-UNIT-NAME TOD)) \\
\>\>\>     WHEN) \\
\>\>  ((FRAME *THAT) \\
\>\> \hspace{.01in} (ROOT THAT) (TYPE PRONOUN)) \\
\>\>  WHEN)

\end{tabbing}
\begin{tabbing}

\hspace{.05in} \= \\

\>{\bf Result1:} Mornings and that are out. \\ \\

\>(\=(FRAME *RESPOND) \\
\>\> (DEGREE NORMAL) \\
\>\> (TYPE NEGATIVE) \\
\>\> (WHEN (*MUL\=TIPLE* \\
\>\>           \>(\=(FRAME *SIMPLE-TIME) \\
\>\>           \>\>(TIME-OF-DAY MORNING) \\
\>\>\>\>        (NUMBER PLURAL) \\
\>\>\>\>        (SIMPLE-UNIT-NAME TOD)) \\
\>\>\>        ((FRAME *THAT) \\
\>\>\>\>       (ROOT THAT) \\
\>\>\>\>       (TYPE PRONOUN))))) \\ 

\end{tabbing}

\end{footnotesize}
\caption{\label{rep-hyps1} {\bf Alternative Repair Hypothesis 1}}
\end{figure}

\begin{figure}[htb]
\begin{footnotesize}
\begin{tabbing}
\hspace{.05in} \= \\

\>{\bf Hypothesis2:} \\ \\

\> (MY-COMB \\
\> \hspace{.05in} \= (\= (TIME-OF-DAY MORNING) \\
\>\>\> (NUMBER PLURAL) \\
\>\>\> (FRAME *SIMPLE-TIME) \\
\>\>\> (SIMPLE-UNIT-NAME TOD)) \\
\>\>((FRAME *RESPOND) \\
\>\>\> (DEGREE NORMAL) \\
\>\>\> (TYPE NEGATIVE)) \\
\>\>  ??)

\end{tabbing}

\begin{tabbing}

\hspace{.05in} \= \\

\>{\bf Result2:} Mornings. \\ \\

\>(\=(FRAME *SIMPLE-TIME) \\
\>\> (TIME-OF-DAY MORNING) \\
\>\>(NUMBER PLURAL) \\
\>\>(SIMPLE-UNIT-NAME TOD)) \\ 

\end{tabbing}
\end{footnotesize}
\caption{\label{rep-hyps2} {\bf Alternative Repair Hypothesis 2}}
\end{figure}

Since a set of alternative meaning representation hypotheses are
constructed during the Combination stage, the result 
is similar to an ambiguous parse.  See Figure \ref{rep-hyps1}
and Figure \ref{rep-hyps2}
for two alternative repair hypotheses produced during the Combination stage
for the example in Figure \ref{parse-example}.
The result of each of the hypotheses is an alternative representation for the
sentence. The first hypothesis, displayed in Figure \ref{rep-hyps1}, 
corresponds to the interpretation, ``Mornings
and that are out.''  The problem with this hypothesis is that it includes
the chunk ``that'', which in this case should be left out.

In the second hypothesis, displayed in Figure \ref{rep-hyps2},
the repair module attempts
to insert the rejection chunk into the time expression chunk, the
opposite of the ideal order.  No slot could be found in the
time expression chunk in which to insert the rejection expression chunk.
In this case, the slot remains uninstantiated and the largest chunk, in this
case the time expression chunk, is returned.  This hypotheses produces a
feature structure that is indeed a portion of the correct structure, though 
not the complete structure.

\subsection{Applying the Genetic Programming Paradigm to Repair}

There are five steps involved in applying the genetic programming paradigm
to a particular problem: determining a set of terminals,
determining a set of functions, determining a fitness measure,
determining the parameters and variables to control the run, and
determining the method for deciding when to stop the evolution process.
The first two constrain the range of repairs that the Repair 
process is capable of making.  The fitness measure determines how
alternative repair hypotheses are ranked, and thus whether it is possible
that the search will converge on the correct hypothesis rather than on
a sub-optimal competing hypothesis.  The last two factors determine how
quickly it will converge and how long it is given to converge.

The set of terminals for this problem is most naturally a chunk from the
parser.  Each operation involved in the repair process takes chunks as input
and returns an augmented chunk as output.  
The single operator, called {\tt my-comb}, takes two chunks as input.
It inserts the second chunk into a slot in the first
chunk.  If it is not possible to insert the second
chunk into the first one, it attempts to merge them. If this too is not 
possible, the largest chunk is returned.

The fitness measure is trained from repair examples from a separate
corpus and is discussed in 
more detail below.  The parameters for the run, such as the size of the
population of programs on each generation, are determined experimentally
from the training corpus.

\subsection{Training a Fitness Function}

The purpose of the trained fitness function is to rank the repair hypotheses 
that are produced in each generation.  Since survival of the fittest is the 
key to the evolutionary process, the determination of which hypotheses are 
more fit is absolutely crucial.  Since the purpose of the repair module is
to evolve a hypothesis that generates the ideal meaning representation
structure, hypotheses that produce meaning representation structures closer 
to the ideal representation should be ranked as better than others that 
produce structures that are more different.  Of course, the repair
module does not have access to that ideal structure while it is
searching for the best combination of chunks.  So a fitness function
is trained that must estimate how close the result of a particular
repair hypothesis is to the ideal structure by considering secondary
evidence.

The first step in training a fitness function is to decide which
pieces of information to make available to the fitness function
for it to use in making its decision.  The fitness function, once it is
trained, combines these pieces of information into a single
score that can be used for ranking the hypotheses.  In the current
version of the ranking
function, three pieces of information are given: the number of
operations in the repair hypothesis, the number of frames and atomic slot
fillers in the resulting meaning representation structure,
and the average of the statistical scores for the set of repairs
that were made.  The statistical score of a repair corresponds to the
mutual information between a slot and the type of filler that was inserted
into it.  This statistical information is trained on a training corpus of
meaning representation structures.

Each piece of information provided to the fitness function is represented as a
numerical score.  The number of operations in the repair hypothesis 
is a measure
of how complex the hypothesis is.  The purpose of this score is to allow
the fitness function to prefer simpler solutions.  The number of frames
and atomic slot fillers is a measure of how complete a repair hypothesis
is.  It allows the fitness function to prefer more complete solutions over
less complete ones.  The statistical scores are a rough measure of the
quality of the decisions that were made in formulating the hypothesis, such
as the decision of which slot in one structure to insert another structure
into. 

The fitness function that combines these three pieces of information is
trained over a training corpus of sentences than need repair coupled with
ideal meaning representation structures.  The purpose of the training process 
is to learn a function that can make wise decisions about the trade offs 
between these three different factors.  Sometimes these three 
factors make conflicting predictions about which
hypotheses are better.  For example, a
structure with a large number of frames that was constructed by
making a lot of statistically unlikely decisions may be less good than a
smaller structure made with decisions that were more likely to
be correct.  The factor that only considers the completeness of the solution
would predict that the hypothesis producing the larger structure
is better.  On the other hand, the factor considering only the statistical
predictions would choose the other hypothesis.  Neither factor will be
correct in all circumstances.
Simple repair hypotheses tend to be better in general, but this goal can
conflict with the goal of having a large resulting structure.  The goal of the
training process is to learn a function that can make these trade-offs
successfully.  

The trained fitness function combines the three given numerical scores using 
addition, subtraction, multiplication, and division.  It is trained using a 
genetic programming technique.  A successful fitness function ranks hypotheses
the same way as an ideal fitness function that can compare the resulting
structures with the ideal one.
Before a fitness function can be trained, there must first be training
data.  Appropriate training data for the fitness function is a set of 
ranked lists of scores, e.g., the three scores mentioned above.  Each set
of three scores corresponds to the repair hypothesis it was extracted from.
These sets of scores in the training examples are ranked the way the 
ideal fitness function would rank the associated hypotheses.  The
purpose of the training process is to find a function that 
combines the three scores into a single score such that when the set of
single scores are sorted, the ordering is the same as in the 
training example.  Correctly sorting the sets of scores is
equivalent to ranking the hypotheses themselves.  Therefore, a function that
can successfully sort the scores in the training examples will be 
correspondingly good at ranking repair hypotheses.

\begin{figure*}[t]
\centerline{\psfig{figure=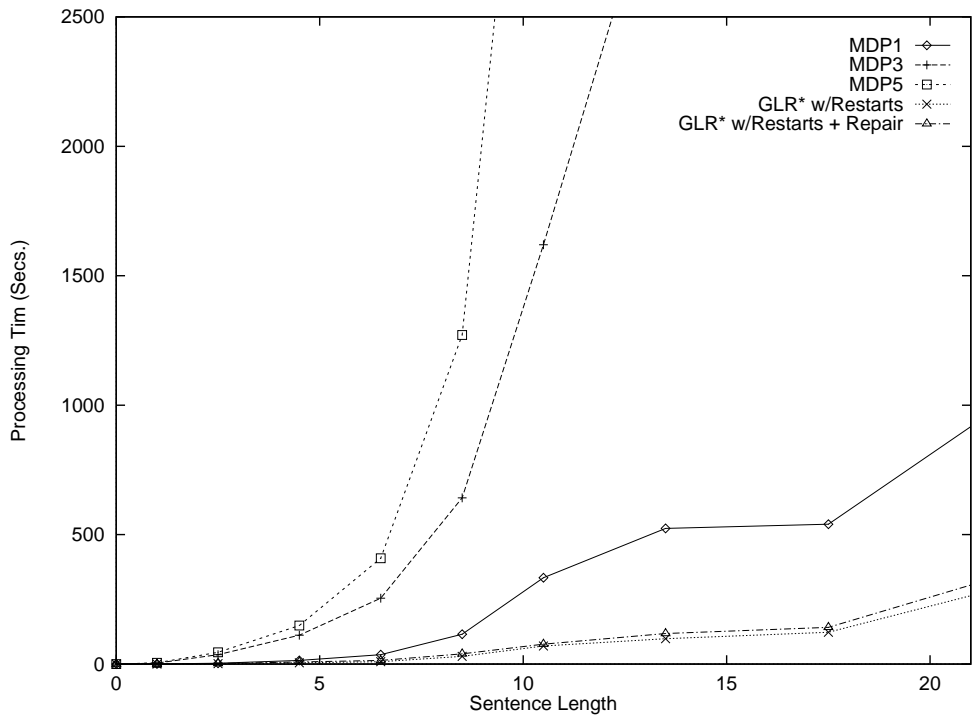,height=3.7in}}
\caption{\bf Parse Times for Alternative Strategies
\label{time}}
\end{figure*}

\section{A Comparative Analysis in a Large Scale Practical Setting}

In order to compare the two stage repair approach with the single stage
MDP approach in a
practical, large-scale scenario, we conducted a comparative evaluation.
As mentioned above, we make use of a version of Lavie's GLR* parser
\cite{lavie} extended to be able to perform both skipping and inserting
which we refer to as LR MDP.
This makes it possible to compare the two stage ROSE approach to MDP 
keeping all other factors constant.

The parser
uses a semantic grammar with approximately 1000 rules which maps the input
sentence onto an interlingua representation (ILT) which represents the meaning 
of the sentence in a language-independent manner.  This ILT is then passed
to a generation component which generates a sentence in the target language
which is then graded by a human judge as Bad, Partial, Okay, or Perfect in 
terms of translation quality.  Partial indicates that the result communicated
part of the content of the original sentence while not containing any
incorrect information.  Okay indicates that the generated sentence communicated
all of the relevant information in the original sentence but not in the ideal
way.  Perfect indicates both that the result communicated the relevant 
information and that it did so in a smooth, high quality manner.
The corpus used in this evaluation contains 500 
sentences from a corpus of
spontaneous scheduling dialogues collected in English.  

In a previous experiment we determined
that the two stage approach performs about two orders of magnitude
faster than LR MDP.  For the purpose of the evaluation presented
in this paper we tested the effect of imposing a maximum
deviation penalty on the minimum distance parser in order to determine how
much flexibility could be allowed before the computational cost would become
unreasonable.  

\begin{figure*}[t]
\centering
\begin{tabular}{|l|c|c|c|c|c|}
\hline 
{\hspace{.25in}} & {NIL} & {Bad} & {Partial} & {Okay} & {Perfect}\\
\hline 
{MDP 1} & {21.4\%} & {3.4\%} & {3.4\%} & {18.4\%} & {53.4\%}\\
\hline 
{MDP 3} & {16.2\%} & {4.2\%} & {5.0\%} & {19.6\%} & {55.0\%}\\
\hline 
{MDP 5} & {8.4\%} & {8.2\%} & {6.0\%} & {21.0\%} & {56.4\%}\\
\hline 
{GLR* with Restarts} & {9.2\%} & {6.4\%} & {12.8\%} & {19.4\%} & {52.2\%}\\
\hline 
{GLR* with Restarts + Repair} & {0.4\%} & {9.6\%} & {11.8\%} & {23.4\%} & {54.8\%}\\
\hline 
\end{tabular}
\caption{Translation Quality of Alternative Strategies}
\label{res}
\end{figure*}

A full, unconstrained implementation of MDP can find an analysis
for any sentence using a combination of insertions, deletions, and 
transpositions.  However, in order to
make it viable to test the MDP approach in a system as large as the one which
provides the context for this work, we make use of a more restricted version
of MDP. While the full MDP algorithm
allows insertions, deletions, and transpositions, our more constrained
version of MDP allows only insertions and deletions.  Although this
still allows the MDP parser to repair any sentence, in some cases the result 
will not be as complete as it would have been with the unconstrained version of
MDP or with the two stage repair process.  Additionally, with a lexicon on the
order of 1000 lexical items, it is not practical to do insertions on the level
of the lexical items themselves.  Instead, we allow only non-terminals to
be inserted.  An insertion penalty
equivalent to the minimum number of words it would take to generate a given
non-terminal is assigned to a parse for each inserted non-terminal.

In order to test the effect of imposing a maximum deviation penalty,
we used a parameterized version of LR MDP,  where the 
deviation penalty of a parse is the total number of words skipped plus
the parse's associated insertion penalty as described above.

The avenues of exploration made available here are far from exhaustive.  
Substitutions and transpositions are not allowed in this version of the
parser, nor is it possible to set 
a separate maximum penalty for skipping and for inserting.  Additionally,
insertions and deletions are weighted equally, where some researchers
have weighted them differently \cite{hipp}.
These and other possibilities are left for future inquiry.

\section{Evaluation}

The LR MDP parser was run over the corpus at three different
flexibility settings.  The first setting, {\bf MDP 1}, is Minimum
Distance Parsing with maximum deviation penalty of 1.  Similarly, 
{\bf MDP 3} and {\bf MDP 5} are MDP with maximum devaition penalty of 
3 and 5 respectively.
We also ran the version of GLR* where only initial segments can
be skipped which we refer to as {\bf GLR* with Restarts}.  Thus, while the parser
can restart from each word in the sentence, analyses produced are always
for contiguous segments of the sentence.  We ran {\bf GLR* with Restarts}
both with and without repair.  Timings for all five of
these iterations over the corpus are displayed in Figure \ref{time}.
Notice that {\bf GLR* with Restarts} is significantly faster than even
{\bf MDP 1}.  And since the repair stage is run only for sentences that
the repair module determines need repair, and since the repair process
takes only seconds on average to run, no significant difference in time can 
be seen in this graph between the case with repair and the case without repair.

The translation quality ratings for the five different iterations over
the corpus are found in Figure \ref{res}.  Predictably, {\bf MDP 5}
is an improvement over {\bf MDP 1} and {\bf MDP 3}, 
with an associated significant cost
in run time.  Also, not surprisingly, the very restricted {\bf GLR* with
Restarts}, while faster than either of the other two, has a
correspondingly lower associated translation quality.  However, {\bf
GLR* with Restarts + Repair} outperforms the other methods in terms of total 
number of acceptable translations, while not being significantly slower 
than {\bf GLR* with Restarts} without repair.  Though these results display
certain trends in the performance of these alternative approaches, the
differences in general are very small.  For example, the difference in number
of acceptable translations between {\bf MDP 5} and {\bf GLR* with restarts + 
repair} is only about 1\%.  The largest difference between the two is that 
{\bf GLR* with
restarts + repair} has about 7\% more sentences with translation quality of
Partial or better, indicating that {\bf GLR* with restarts + repair} produces
analyses that are useful for furthering the conversation between the two
speakers using the system 7\% more often than {\bf MDP 5}.

While we have no doubt that increasingly more flexible versions of MDP
would perform better than {\bf MDP 5}, we have already
demonstrated that even {\bf MDP 5} is impractical in terms of its run-time
performance.  Thus we conclude that the two stage ROSE approach, even with
a very limited flexibility parser, is a superior choice.  We believe that
by increasing the flexibility of the parser to include very limited skipping
in addition to restarts would increase the performance of this two stage
approach without incurring a significant increase in run time performance.
Determining exactly how much skipping is ideal is a direction for future
research.

\section{Conclusions}

In this paper we addressed the issue of how to efficiently handle the problem 
of extragrammaticality in a large-scale spontaneous spoken language system.
We argue that even though Minimum Distance Parsing offers a theoretically 
attractive solution to the problem of extragrammaticality, it is 
computationally infeasible in large scale practical applications.  
Our analysis demonstrates that the ROSE approach, consisting of a skipping 
parser with limited flexibility coupled 
with a completely automatic post-processing repair module, performs
significantly faster than even a version of MDP limited only to skipping and 
inserting and constrained to a maximum deviation penalty of 5, while producing
analyses of superior quality.


\begin{thebibliography}{}

\bibitem[\protect\citename{Abney}1996]{abney}
Abney, S.
\newblock 1996.
\newblock Partial parsing via finite-state cascades.
\newblock In {\em Proceedings of the Eight European Summer School In Logic,
  Language and Information, Prague, Czech Republic}.

\bibitem[\protect\citename{Danieli and Gerbino}1995]{danieli}
Danieli, M. and E.~Gerbino.
\newblock 1995.
\newblock Metrics for evaluating dialogue strategies in a spoken language
  system.
\newblock In {\em Working Notes of the AAAI Spring Symposium on Empirical
  Methods in Discourse Interpretation and Generation}.

\bibitem[\protect\citename{Ehrlich and Hanrieder}1996]{uteger}
Ehrlich, U. and G.~Hanrieder.
\newblock 1996.
\newblock Robust speech parsing.
\newblock In {\em Proceedings of the Eight European Summer School In Logic,
  Language and Information, Prague, Czech Republic}.

\bibitem[\protect\citename{Federici, Montemagni, and Pirrelli}1996]{fedmonpir}
Federici, S., S.~Montemagni, and V.~Pirrelli.
\newblock 1996.
\newblock Shallow parsing and text chunking: a view on underspecification in
  syntax.
\newblock In {\em Proceedings of the Eight European Summer School In Logic,
  Language and Information, Prague, Czech Republic}.

\bibitem[\protect\citename{Hipp}1992]{hipp}
Hipp, D.~R.
\newblock 1992.
\newblock {\em Design and Development of Spoken Natural-Language Dialog Parsing
  Systems}.
\newblock {Ph.D.} thesis, Dept.\ of Computer Science, Duke University.

\bibitem[\protect\citename{Koza}1992]{gen2}
Koza, J.
\newblock 1992.
\newblock {\em Genetic Programming: On the Programming of Computers by Means of
  Natural Selection}.
\newblock MIT Press.

\bibitem[\protect\citename{Koza}1994]{gen3}
Koza, J.
\newblock 1994.
\newblock {\em Genetic Programming {II}}.
\newblock MIT Press.

\bibitem[\protect\citename{Lavie}1995]{lavie}
Lavie, A.
\newblock 1995.
\newblock {\em A Grammar Based Robust Parser For Spontaneous Speech}.
\newblock {Ph.D.} thesis, School of Computer Science, Carnegie Mellon
  University.

\bibitem[\protect\citename{Lavie \bgroup et al.\egroup }1996]{janus96}
Lavie, A., D.~Gates, M.~Gavalda, L.~Mayfield, and A.~Waibel~L. Levin.
\newblock 1996.
\newblock Multi-lingual translation of spontaneously spoken language in a
  limited domain.
\newblock In {\em Proceedings of COLING 96, Kopenhagen}.

\bibitem[\protect\citename{Lavie and Tomita}1993]{lt}
Lavie, A. and M.~Tomita.
\newblock 1993.
\newblock {GLR*} - an efficient noise-skipping parsing algorithm for context
  free grammars.
\newblock In {\em Proceedings of the Third International Workshop on Parsing
  Technologies}.

\bibitem[\protect\citename{Lehman}1989]{lehman}
Lehman, J.~F.
\newblock 1989.
\newblock {\em Adaptive Parsing: Self-Extending Natural Language Interfaces}.
\newblock {Ph.D.} thesis, School of Computer Science, Carnegie Mellon
  University.
\newblock CMU-CS-89-191.

\bibitem[\protect\citename{Nord}1996]{vanord}
Nord, G.~Van.
\newblock 1996.
\newblock Robist parsing with the head-corner parser.
\newblock In {\em Proceedings of the Eight European Summer School In Logic,
  Language and Information, Prague, Czech Republic}.

\bibitem[\protect\citename{Srinivas \bgroup et al.\egroup }1996]{srin}
Srinivas, B., C.~Doran, B.~Hockey, and A.~Joshi.
\newblock 1996.
\newblock An approach to robust partial parsing and evaluation metrics.
\newblock In {\em Proceedings of the Eight European Summer School In Logic,
  Language and Information, Prague, Czech Republic}.

\bibitem[\protect\citename{Woszcyna \bgroup et al.\egroup }1994]{janus2}
Woszcyna, M., N.~Aoki-Waibel, F.~D. Buo, N.~Coccaro, K.~Horiguchi, T.~Kemp,
  A.~Lavie, A.~McNair, T.~Polzin, I.~Rogina, C.~P. Ros\'{e}, T.~Schultz,
  B.~Suhm, M.~Tomita, and A.~Waibel.
\newblock 1994.
\newblock {JANUS} 93: Towards spontaneous speech translation.
\newblock In {\em Proceedings of the International Conference on Acoustics,
  Speech, and Signal Processing}.

\bibitem[\protect\citename{Woszcyna \bgroup et al.\egroup }1993]{janus}
Woszcyna, M., N.~Coccaro, A.~Eisele, A.~Lavie, A.~McNair, T.~Polzin, I.~Rogina,
  C.~P. Ros\'{e}, T.~Sloboda, M.~Tomita, J.~Tsutsumi, N.~Waibel, A.~Waibel, and
  W.~Ward.
\newblock 1993.
\newblock Recent advances in {JANUS:} a speech translation system.
\newblock In {\em Proceedings of the {ARPA} Human Languages Technology
  Workshop}.

\end{thebibliography}
\end{document}